\documentclass{article}

\usepackage[preprint, nonatbib]{neurips_2024}

\usepackage[utf8]{inputenc} % allow utf-8 input
\usepackage[T1]{fontenc}    % use 8-bit T1 fonts
\usepackage{hyperref}       % hyperlinks
\usepackage{url}            % simple URL typesetting
\usepackage{booktabs}       % professional-quality tables
\usepackage{amsfonts}       % blackboard math symbols
\usepackage{nicefrac}       % compact symbols for 1/2, etc.
\usepackage{microtype}      % microtypography
\usepackage{xcolor}         % colors
\usepackage{listings}
\usepackage{graphicx}
\usepackage{biblatex}

\title{2BP: 2-Stage Backpropagation}

\author{
  Christopher Rae \\
  \texttt{crae@ed.ac.uk} \\
  EPCC, The University of Edinburgh\\
  \AND
  Joseph Lee \\
  \texttt{j.lee@epcc.ed.ac.uk} \\
  EPCC, The University of Edinburgh\\
  \And
  James Richings \\
  \texttt{j.richings@epcc.ed.ac.uk} \\
  EPCC, The University of Edinburgh\\
}

\begin{document}

\maketitle

\begin{abstract} %-------------------------------------------------------------------------------------
  As Deep Neural Networks (DNNs) grow in size and complexity, they often exceed the memory capacity of a single accelerator, necessitating the sharding of model parameters across multiple accelerators. Pipeline parallelism is a commonly used sharding strategy for training large DNNs. However, current implementations of pipeline parallelism are being unintentionally bottlenecked by the automatic differentiation tools provided by ML frameworks. This paper introduces 2-stage backpropagation (2BP). By splitting the backward propagation step into two separate stages, we can reduce idle compute time. We tested 2BP on various model architectures and pipelining schedules, achieving increases in throughput in all cases. Using 2BP, we were able to achieve a 1.70x increase in throughput compared to traditional methods when training a LLaMa-like transformer with 7 billion parameters across 4 GPUs.
\end{abstract}

\section{Introduction} %-------------------------------------------------------------------------------------
In the era of big data and artificial intelligence, deep neural networks (DNNs) have emerged as powerful tools for a wide range of applications, from natural language processing to genomic sequencing. As these models grow in size and complexity they can no longer fit within the memory of a single accelerator, leading to a variety of novel techniques being introduced, including 3D parallelism\cite{megatronlm, deepspeed, colossalai}, Zero Redundancy Optimizer(ZeRO) \cite{zero} and parameter offloading to host memory or NVMe storage \cite{patrickstar, elixir, colossalai, zero_offload}. 3D parallelism refers to the three main strategies for parallelising the computation of DNNs: firstly, data parallelism, which refers to the technique of dividing a dataset across multiple accelerators each with their own copies of the model; pipeline parallelism, which distributes model layers across multiple accelerators; and lastly tensor parallelism, which splits individual model layers across multiple accelerators. 
While data parallelism does not directly address the issue of increasing model sizes, it enhances processing speed and can be used to reduce the memory required for activations by reducing the local batch size for each accelerator.
The last two forms of parallelism fall under the umbrella of model parallelism, and are often necessitated by models with sizes exceeding the capacity of a single accelerator. In particular, tensor parallelism is particularly effective only for extremely large models due to large amount communication overhead, which results in inefficient resource utilization when applied to smaller models.

Focusing on pipeline parallelism, naive implementations often result in a long proportion of idle compute time, which is caused by computational dependencies. The ratio of idle compute to compute is often referred to as the \textit{bubble ratio}. To reduce the bubble ratio, a range of pipeline parallelism algorithms have been developed \cite{gpipe, dapple, chimera, pipedream, pipemare, wpipe, hanayo}. Pipeline parallel algorithms can be split into 2 categories: synchronous and asynchronous. Synchronous algorithms require a flush at the end of each training step, in order to keep all of the accelerators training on the same version of the model weights. Asynchronous algorithms do not preform this flush, often resulting in a smaller bubble ratio at the cost of not converging as well as synchronous algorithms \cite{pipedream, pipemare, wpipe}.

Deep neural networks are generally composed of a sequence of layers. The output of each layer $l$ is a function of the parameters (e.g. weights and biases) within that layer $w_l$ and the output of the previous layer $z_{l-1}$, such that $z_l = f_l(z_{l-1}, w_l)$. The basic training of DNNs involves three primary steps: forward propagation, used to calculate the intermediate activations, backward propagation, where we calculate the gradients of the model's parameters and finally, the optimizer step which updates the model's parameters using the calculated gradient. In practice, the backward propagation step is typically abstracted away by the machine learning framework, and not directly programmed by the user. ML frameworks such as PyTorch \cite{pytorch} and TensorFlow \cite{Tensorflow} use reverse mode automatic differentiation to calculate the gradients, which often involves tracing the computation graph produced by the forward pass at runtime and calculating the derivatives by starting at the end of the graph and propagating to the beginning. The intuition behind reverse mode automatic differentiation stems from the chain rule. As we backward propagate through each layer in our neural network, two calculations take place: the partial derivative of the loss with respect of the parameters $\frac{\partial L}{\partial w_{l}}$ and the partial derivative of the loss with respect to the input of the layer $\frac{\partial L}{\partial z_{l-1}}$. Both operations depend on a combination of the activations calculated in the forward propagation and the derivative of the loss with respect to the output $\frac{\partial L}{\partial z_l}$. Both of these calculations are handled by the ML framework under a single predefined backward function. This approach works well in a single accelerator setting as it is more memory efficient not having to store the intermediate derivatives $\frac{\partial L}{\partial z_l}$ for later computation, but as we scale our model and pipeline parallelism becomes necessary, the calculation of $\frac{\partial L}{\partial w_{l}}$ is not immediately required, and can be reordered such that it is calculated only after the backward propagation has started on the preceding accelerator. The objective of this work is to apply this splitting and reordering of the backward propagation step, and measure the performance gained for a variety of model architectures.

\section{Background \& related work} %-------------------------------------------------------------------------------------
A wide range of pipeline parallelism schedules have been introduced and studied. The concept of a pipelining schedule was first introduced with GPipe \cite{gpipe}, a pipeline parallelism library developed by Google. GPipe introduced the innovative idea of splitting mini-batches further into micro-batches, allowing different accelerators to train on multiple micro-batches in parallel. This approach effectively decreases the bubble ratio by allowing for overlapped compute between accelerators, thereby enhancing the efficiency of the training process. 

The next prominent synchronous pipelining schedule to emerge was 1F1B (1 forward, 1 backward), sometimes referred to as PipeDream-Flush. This schedule was originally proposed by PipeDream \cite{pipedream} as an asynchronous pipeline schedule, but was later adapted for use in Megatron-LM \cite{megatronlm} and DAPPLE \cite{dapple} as a synchronous pipelining schedule. Megatron-LM is a research-oriented framework designed for large language model (LLM) training at scale, while DAPPLE is a synchronous training framework that combines data parallelism and pipeline parallelism for training large DNNs. It is also important to note that DeepSpeed \cite{deepspeed} independently developed a pipeline scheduling algorithm that is equivalent to 1F1B. To further reduce the bubble ratio when using 1F1B, an interleaved pipelining schedule \cite{megatronlm} can be used. We can use this interleaved schedule to decrease the idle compute at the cost of an increase in communication.

Chimera \cite{chimera} is another noteworthy bidirectional pipeline scheduling algorithm. This method involves storing two copies of the model across the accelerators, which results in doubling the memory consumed by the model's weights. Despite this increase in memory usage and the need for a more complex communication scheme, Chimera significantly reduces the bubble ratio by up to 50\%, making it a highly efficient scheduling strategy. 

There are a handful of libraries that have implemented tools to utilise pipeline parallelism at runtime. FairScale \cite{FairScale2021} provides an implementation of GPipe for PyTorch sequential modules. DeepSpeed offers their own implementation of 1F1B also for PyTorch sequential modules. Megatron-LM allows users to use 1F1B or 1F1B interleaved to train LLMs. Colossal-AI \cite{colossalai} (much like Megatron-LM) offers 1F1B or 1F1B interleaved support for a wide range of transformers. 

At a similar time as this work is produced, Zero Bubble Pipeline Parallelism \cite{zbpp} was introduced, which also separates the backward propagation step into 2 separate stages. The authors adds support for 2BP (described in later Section~\ref{imp details}) on top of the linear layers in Megatron-LM as well as implementing a custom pipelining schedules that uses 2BP to further reduce bubble time to theoretically zero. Our work evaluates the performance of 2BP on a wide range of model architectures and evaluates the scaling performance of the strategy.

\section{2-Stage Backpropagation} \label{imp details} %-------------------------------------------------------------------------------------
\subsection{Definitions}

For each layer during back propagation, we define the computation of the gradient w.r.t to the output of the preceding layer $\frac{\partial L}{\partial z_{l-1}}$ as `backward-p1' and computation of the gradient w.r.t to the layer parameters $\frac{\partial L}{\partial w_{l}}$ as `backward-p2'. We describe 2BP as the process of splitting up the backward pass into backward-p1 and backward-p2, and delaying the computation of backward-p2 in order to maximise accelerator utilization. 2BP can be applied on top of any pipelining schedule, including Gpipe and 1F1B. Figure~\ref{pipe schedules} displays the effects of different pipelining schedules, with and without the use of 2BP.

\subsection{Implementation}
Our implementation is built on top of PyTorch. However, we do not use PyTorch's automatic differentiation engine (\texttt{torch.autograd}). This granted us greater flexibility when implementing 2BP. We replaced \texttt{torch.autograd} by manually implementing the backward pass operation for all of the modules employed within the models we chose to benchmark. Each module has a forward and a backward-p1 function; if that module contains parameters (equivalent to \texttt{torch.nn.Parameter}) then it also has a backward-p2 function. We can simulate the behaviour of \texttt{torch.autograd} by calling backward-p2 directly after a backward-p1 call as we backward propagate through the compute graph, instead of delaying the computation.

\begin{figure}[h]
  \centering
  \includegraphics[width=\linewidth]{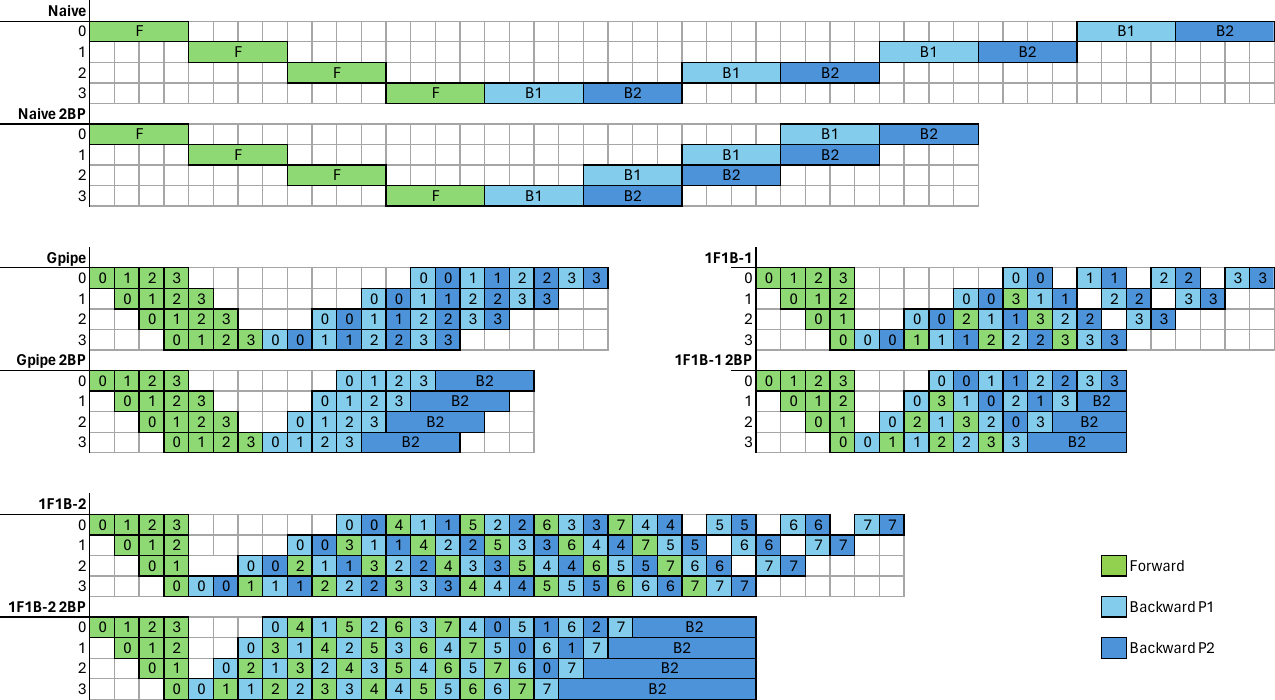}
  \caption{Pipelining schedules (Naive, Gpipe, 1F1B-1, 1F1B-2) with and without 2BP. This figure assumes that the time taken to compute the forward, backward-p1 and backward-p2 passes are equal.}
  \label{pipe schedules}
\end{figure}

We test the effects of 2BP on 4 model architectures across 4 pipelining schedules. The first model chosen was a transformer architecture \cite{attention} with 7 billion parameters (Transformer-7b). This model is based on the LLaMa \cite{llama} and PaLM \cite{palm} architectures (steps including rotary embedding \cite{roformer}, SwiGLU MLP \cite{swiglu}, RMSNorm \cite{rmsnorm}, no linear bias). Transformer-7b has a context length of 1024 and a model dimension of 4096. The second chosen model is another well-known bidirectional transformer model, BERT-Large \cite{bert}. To expand the test set beyond transformer models, a relatively new architecture, Mamba-1.4B \cite{mamba} was chosen. Lastly, to evaluate the performance on an architecture with a non-uniform computational graph, we have chosen ResNet-152 \cite{resnet}, a convolutional neural network (CNN) model. A model with a non-uniform compute graph refers to a model who's activations do not share a constant shape throughout the model, transformers are an example of a model with a uniform compute graph as their activations keep a constant shape at every block. By selecting these four diverse models, we ensure a comprehensive assessment of 2BP across various architectural designs and computational demands. 

The 4 pipelining schedules we test are Gpipe, 1F1B-1, 1F1B-2 and a naive approach with no scheduling algorithm and maximum bubble time. 1F1B is split into 2 categories: 1F1B-1 which indicates that the mini-batch is split into $N$ micro-batches where $N$ is the number of pipeline processes (also equal to the number of accelerators), and 1F1B-2 which indicates that the mini-batch is split into $2N$ micro-batches. 

\begin{figure}[h]
  \centering
  \includegraphics[width=2cm]{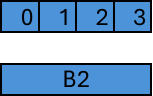}
  \caption{Combining each microbatch's backward-p2 into a single operation.}
  \label{concat b}
\end{figure}

For our GPipe implementation, we delay the computation of backward-p2 until all micro-batches have finished both forward and backward-p1 steps. We then concatenate the activations and intermediate derivatives of all micro-batches over the batch dimensions, meaning we only have to call backward-p2 once rather than $N$ times as shown in Figure \ref{concat b}. For both 1F1B implementations  (without 2BP), every accelerator (except the last) has some idle time after their backward-p1 calls. After the backward-p1 operation has been called for a micro-batch, all the necessary intermediates have been calculated and stored in memory for use in the computation of backward-p2 for that micro-batch. This allows us to fill that idle time between backward-p1 calls with backward-p2 calls (see Figure \ref{pipe schedules} for a visual representation). Once all the backward-p1 steps have been called, we concatenate the remaining forward and backward-p1 intermediates over the batch dimension and compute backward-p2 the same way as was done with GPipe. When working with models with non-uniform compute graphs (such as CNNs), executing backward-p2 steps during idle time may not be optimal, since the backward-p2 step may take longer than the original idle time, resulting in excess idle time on a different accelerator. 

Table~\ref{bubble time} shows the theoretical bubble ratios as functions of $N$ where $N$ is the number of pipeline processes. Bubble ratio is defined as the fraction of total runtime which is idle, i.e. idle time divided by the total runtime. These functions assume that the forward, backward-p1 and backward-p2 functions take equal lengths of time, which does not hold true in practice but acts as a good estimate for comparing the performance scaling for the different schedules as the number of processes varies. The throughput gain is the speedup of the total runtime from using and not using 2BP, and can be obtained from $\frac{1- b}{1-a}$, where $b$ is the bubble ratio with 2BP and $a$ is the bubble ratio without.

\begin{table}[h]
  \centering
  \caption{Bubble ratios and throughput gains for each pipelining schedule. Equal time for forward, backward-p1, and backward-p2 assumed.}
  \label{bubble time}

  \begin{tabular}{c c c c}
   \toprule
   Pipelining Schedule & Bubble Ratio &  2BP Bubble Ratio & Throughput Gain \\
   \midrule %(1-b)/(1-a)
    Naive   & $\frac{N-1}{N}$    &  $\frac{2(N-1)}{2N+1}$     & $\frac{3N}{2N+1}$ \\\\
    GPipe   & $\frac{N-1}{2N-1}$ & $\frac{2(N-1)}{2(N-1)+3N}$ & $\frac{3(2N-1)}{2(N-1) + 3N}$ \\\\
    1F1B-1  & $\frac{N-1}{2N-1}$ & $\frac{N-1}{N-1 + 3N}$     & $\frac{3(2N-1)}{N-1 + 3N}$  \\\\
    1F1B-2  & $\frac{N-1}{3N-1}$ &  $\frac{N-1}{N-1 + 6N}$    & $\frac{3(3N-1)}{N-1 + 6N}$\\
   \bottomrule
  \end{tabular}
\end{table}

We also used \texttt{torch.jit.script} to compile the backward-p1 operations for both softmax and RMSNorm, resulting in significant speed up for each. We do not include this pre-training compilation step in our timing measurements, since this is required whether 2BP is in place or not. The data collected came from training on randomly generated data. This is done instead of using an actual dataset since from our experience with the system, dataloading can be a significant bottleneck and optimising dataloading is beyond the scope of this paper. All of the throughput results were collected from training on randomly generated data samples of size 16384 and averaged over 4 epochs.

\section{Evaluation} %-------------------------------------------------------------------------------------

Our experiments were performed on two systems at EPCC: the Edinburgh International Data Facility (EIDF) GPU-Service \cite{eidf, eidf_gpuservice} and Cirrus \cite{cirrus_hardware}. Most of the results were collected from the GPU nodes on the EIDF GPU-service containing 4 40GB Nvidia A100 connected via SXM4 interconnect, and the Mamba throughput results were collected on nodes with 4 80GB Nvidia A100s connected via SXM4 interconnect. Cirrus supports multi-node GPU jobs, which is particularly useful for scaling tests. Cirrus GPU nodes contains 4 16GB Nvidia V100 GPUs connected via an SXM2 interconnect. For our PyTorch implementation, since autograd is not used, we are able to perform our runs with PyTorch's inference mode activated. PyTorch's distributed library is used for p2p communication with a NCCL \cite{nccl} backend.

Table \ref{benchamrk_table} summarises the model architecture used for our experiments. Although the optimizer \cite{adam, adamw} chosen for each benchmark is not relevant for this paper, the optimizer calculations are taken into account during the throughput measurements. The naive implementation does not use micro-batches but uses a mini-batch of $4 \times$micro-batch size. When training ResNet152, in order to keep the batch normalization \cite{batchnorm} computation equal between pipeline schedule, the naive implementation uses a mini-batch size of 8 and 4 gradient accumulation steps. All models with the exception of ResNet152 distributed the number of blocks equally amongst the 4 GPUs (excluding the embedding blocks and prediction heads where appropriate). ResNet152 contains 50 ResNet bottlenecks which were split in the ratio [10, 14, 14, 12] across the 4 GPUs respectively. ResNet152 also contains some initial convolutions computed by GPU 0 and a classification head computed by GPU 3. The loss is always handled by GPU 3 as it computes the model's output.

\begin{table}[h]
  \centering
  \caption{Model hyperparameters used for benchmarking.}
  \label{benchamrk_table}

  \begin{tabular}{c c c c}
   \toprule
   Model &  Data type &  Micro-Batch size & Optimizer\\
   \midrule
    Mamba-1.4b & fp16  & 2 & AdamW \\
    LLaMa-7b   & fp16  & 1 & Adam  \\
    ResNet152  & fp32  & 8 & SGD   \\
    BERT-Large & fp16  & 2 & Adam  \\
   \bottomrule
  \end{tabular}
\end{table}

\subsection{2BP throughput}

\begin{figure}[h]
  \centering
  \includegraphics[width=12cm]{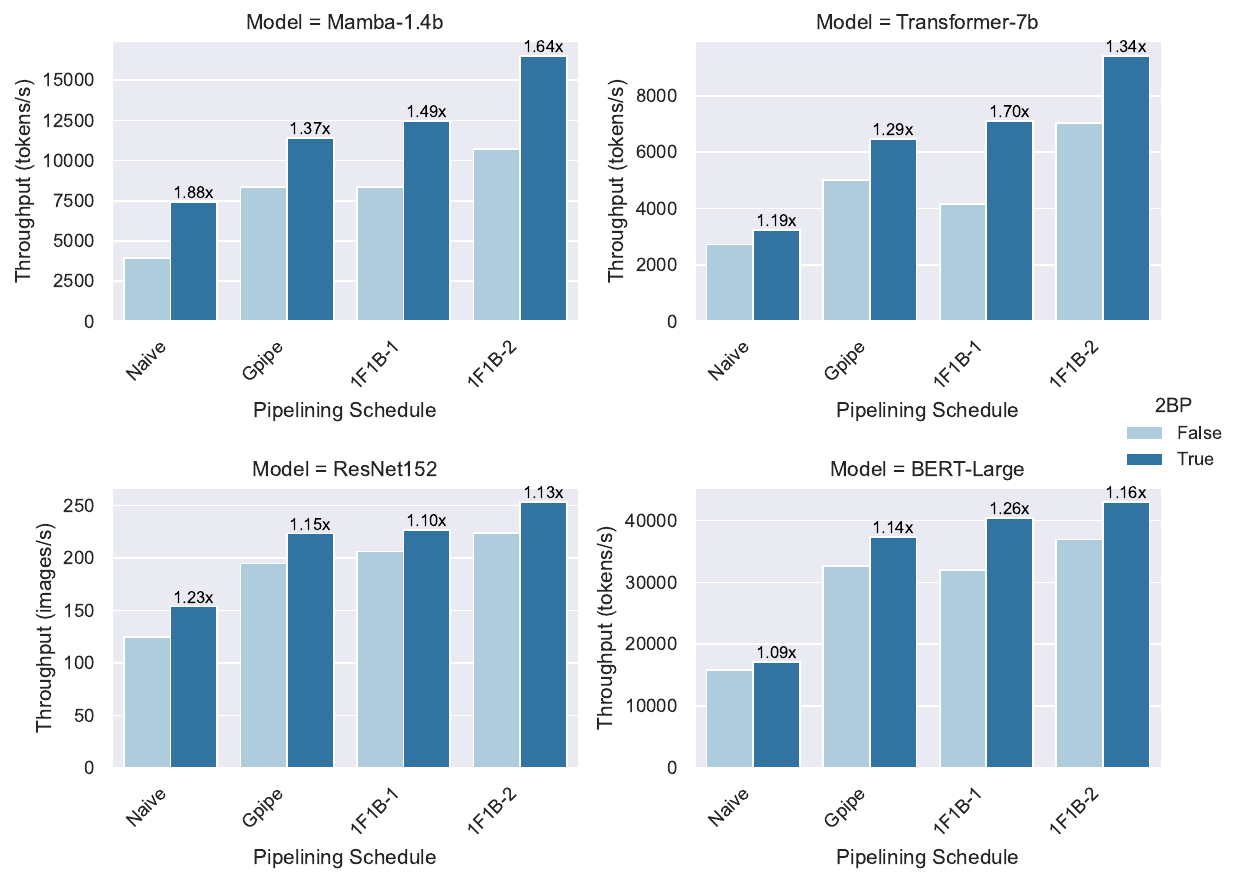}
  \caption{Sample throughput for each model with different pipeline schedules. Light blue bars represent schedule without 2BP, and dark blue represents with 2BP. Numbers above bars represent throughtput gain from using 2BP.}
  \label{throughput}
\end{figure}

Figure~\ref{throughput} summarises the throughput results for different models with different pipeline schedules, with and without 2BP. Speedup can be observed for every case with the use of 2BP, ranging from \textbf{1.10x} for 1F1B-1 on ResNet152 to up to \textbf{1.70x} for 1F1B-1 on Transformer-7b. The observed difference in performance gain for different models may be caused by multiple factors, including the non-uniformity of the ResNet152 model results in unwanted idle compute time, as well as the smaller problem size compared to models like Transformer-7b, this would also explain the smaller performance gain of BERT-Large at \textbf{1.26x}. 

The time taken for the backward-p1 step compared to backward-p2 step is usually uneven, and depends heavily on the model architecture. For example, for 2D batch normalization, the backward-p2 operation is significantly simpler than the backward-p1 operation. Furthermore, some operations, e.g. the scalar dot-product attention, do not require a backward-p2 operation but have a significant backward-p1 operation. This difference in computational complexity between backward-p1 and backward-p2 may cause the variation of performance gain observed between model architectures.

\subsection{2BP memory consumption}

\begin{figure}[h]
  \centering
  \includegraphics[width=12cm]{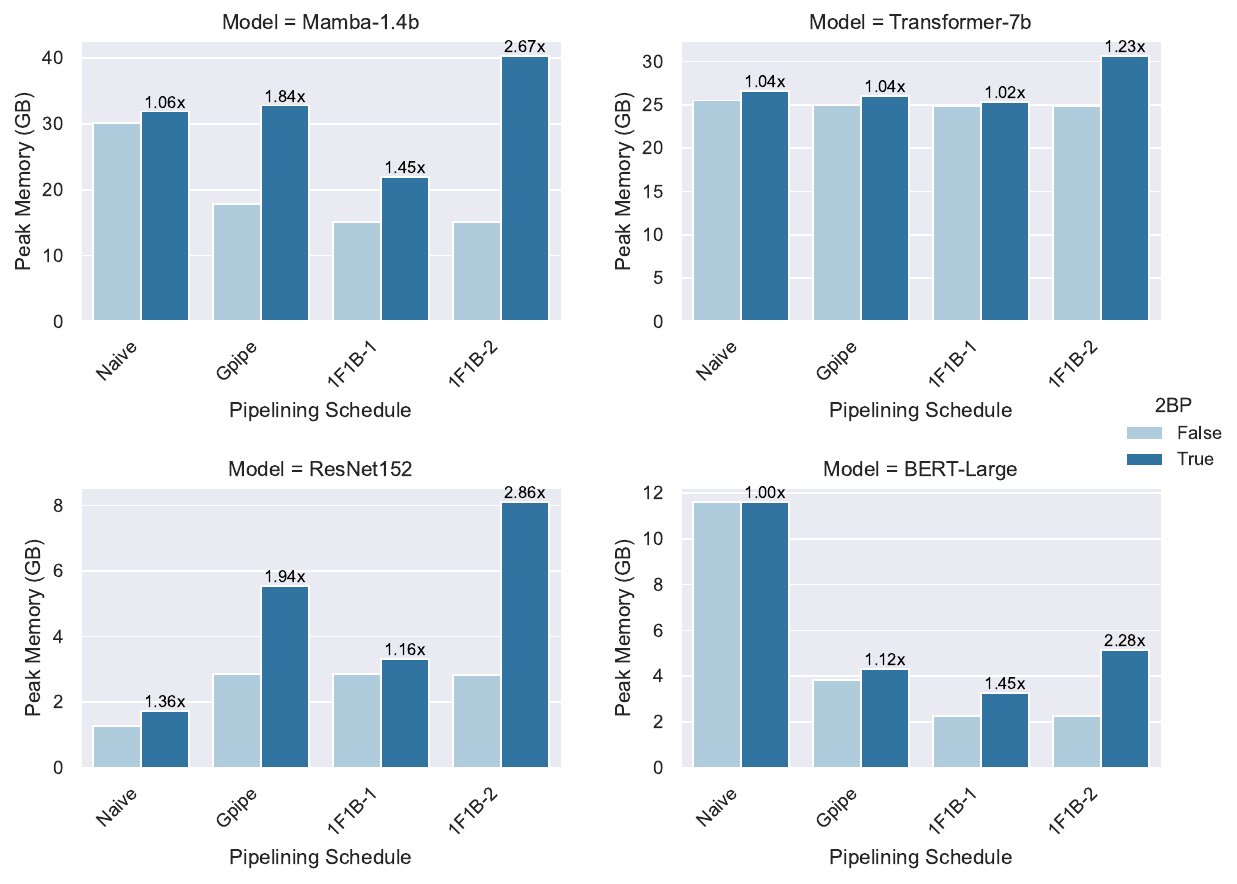}
  \caption{Maximum memory usage across the 4 GPUs. The "peak memory (GB)" is measured by obtaining the peak reserved memory on each GPU and taking then the maximum. Light blue bars represent schedule without 2BP, and dark blue represents with 2BP. Numbers above bars represent the increase in memory from using 2BP.}
  \label{Peak Memory}
\end{figure}

While 2BP is able to reduce idle computation time, it comes at the cost an increase in peak memory usage. Figure~\ref{Peak Memory} summarises the effect on peak GPU memory usage (i.e. the GPU which reserved the maximum memory) for each model and pipeline schedule, with and without 2BP. It can be seen that the increase in memory consumption varies significantly between the model architectures but also the pipeline schedules. In Mamba with the 1F1B-2 pipelining schedule, we see an increase in memory of \textbf{2.67x}; on the other hand, for training Transformer-7b with the 1F1B-1 pipelining schedule, the memory consumption only increases by \textbf{1.02x}.

The increase in memory is a result of 2 different characteristics of 2BP: first the intermediate derivatives $\frac{\partial L}{\partial z_l}$ have to be stored in memory after backward-p1 for reuse in backward-p2. Second, the activations that are needed for the delayed backward-p2 calls are held for longer. The GPU with the largest memory consumption is dependent on the pipelining schedule being used and the uniformity of the models compute graph.

The specific operations used in a model can result in a significant increase in peak memory; for Linear and Convolution layers, both the input activations and output derivatives need to be stored in memory for backward-p2. On the other hand, operations that are purely functional such as ReLU and Scalar Dot-Product Attention release their activations during the backward-p1 calls. 

Comparing the memory footprint for different pipelining schedules, Gpipe requires all process to save the activations of all the micro-batches for the backward pass, resulting in a large memory footprint for each device. For 1F1B-1, without the use of 2BP, GPU 0 will always have the largest activation memory since it has to store the activations for $N$ micro-batches; with 2BP, the GPU with the greatest activation memory is now dependent on the model architecture. Although GPU 0 has the activations of all micro-batches saved in memory, it only ever has to store $1$ micro-batch worth of intermediate derivatives. GPU $N-1$ has to store $N$ micro-batches worth of intermediate derivatives. The peak memory required by activations is dependent on how many activations are released after backward-p1.

\begin{figure}[h]
  \centering
  \includegraphics[width=10cm]{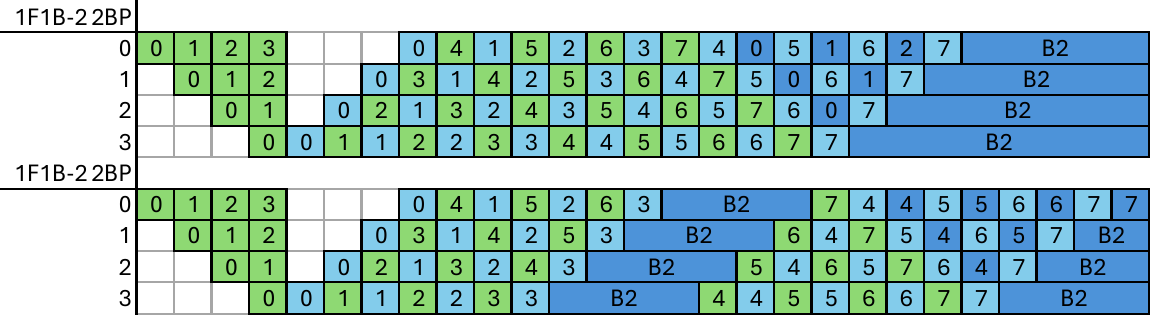}
  \caption{Alternative memory efficient schedule for 1F1B-2 with 2BP}
  \label{Memory Efficient}
\end{figure}

The reason for the large increase in peak memory usage for 1F1B-2 as seen by the is the fact that the majority of the activations and intermediate derivatives need to be held in memory until backward-p2 is called. Although we did not implement this, by calling backward-p2 on some of the micro-batches halfway through the 1F1B-2 training step, as visualised in Figure \ref{Memory Efficient}, it may be possible to recover the peak memory usage closer to that of 1F1B-1. As further work to this study, we would like to investigate how the frequency at which backward-p2 is called during a training step affects performance and memory consumption, especially as we scale to versions of 1F1B with greater numbers of micro-batches e.g. $8N$ micro-batches.

\subsection{Scaling}
In this section we look into the effects of scaling on 2BP. Table \ref{bubble time} shows that as the number of accelerators $N$ increases, the bubble ratio increases. At the same time the performance gain between training with and without 2BP should increase since the rate at which the bubble ratio increases is greater when not using 2BP. Using a BERT-like model, we test the effect of scaling when using 1F1B-1 and 1F1B-2 as they give the greatest throughput. Both implementations use a micro-batch size of 2.

\subsubsection{Fixed model size} \label{section scaling constant}

\begin{figure}[h]
  \centering
  \includegraphics[width=12cm]{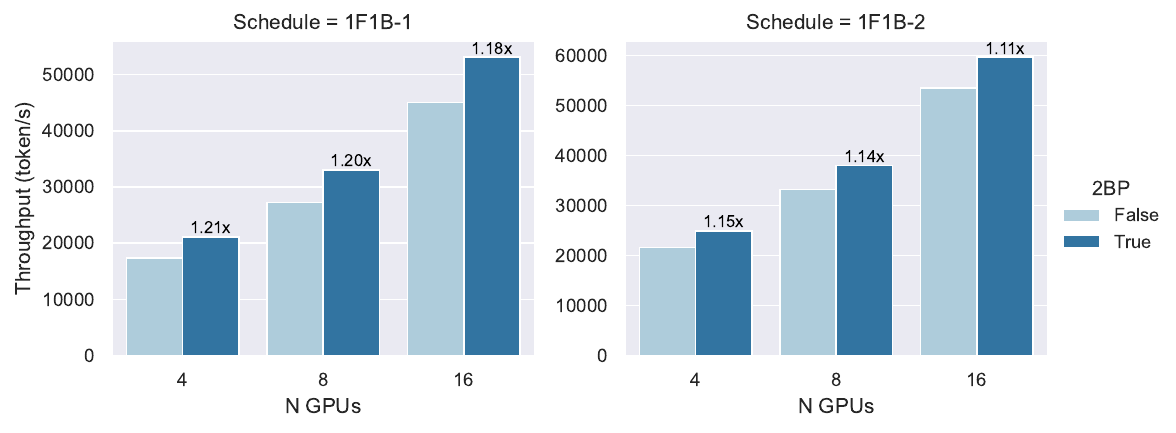}
  \caption{Scaling the number of GPUs with a fixed model size. Light blue bars represent schedule without 2BP, and dark blue represents with 2BP. Numbers above bars represent throughtput gain from using 2BP.}
  \label{scaling constant}
\end{figure}

We use a BERT-like model with 32 blocks to perform a scaling test with fixed model size. Figure \ref{scaling constant} summarises the effect of 2BP with varying number of accelerators. For 1F1B-1, the performance increase from 2BP goes from \textbf{1.21x} for 4 GPUs, to \textbf{1.20x} for 8 GPUs, and \textbf{1.18x} for 16 GPUs. For 1F1B-2, the gain is \textbf{1.15x}, \textbf{1.14x} and \textbf{1.11x} respectively. Even though the theoretical prediction from Table~\ref{bubble time} suggests that the throughput gain should increase as the number of accelerators increase, the formula did not take into account the increase in communication required, especially when going above 4 GPUs (on our system) requires inter-node communication. 

\subsubsection{Variable model size}

\begin{figure}[h]
  \centering
  \includegraphics[width=12cm]{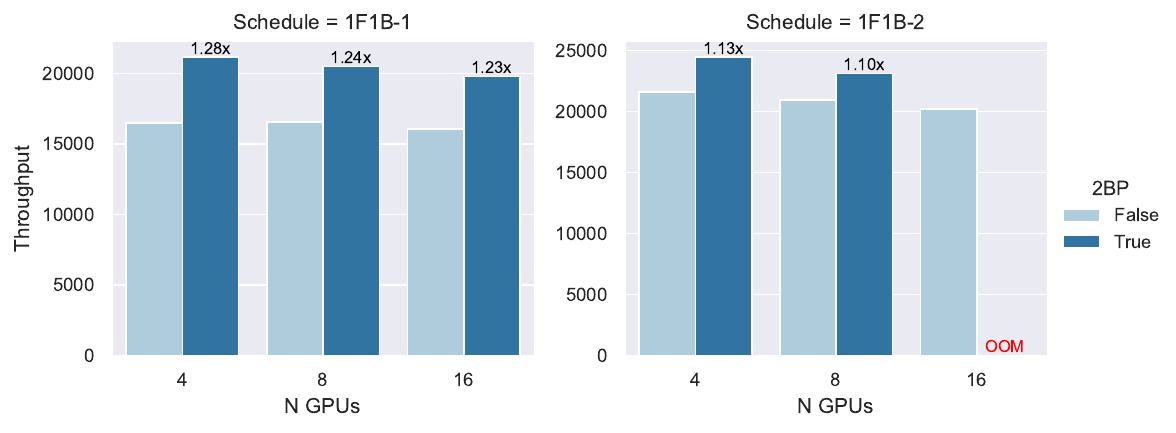}
  \caption{Scaling the number of GPUs with a variable global model size. Light blue bars represent schedule without 2BP, and dark blue represents with 2BP. Numbers above bars represent throughtput gain from using 2BP.}
  \label{scaling variable}
\end{figure}

Here we test the performance as we scale the model size with the number of processors; each accelerator computes 8 BERT-like blocks. From Figure~\ref{scaling variable}, we see that for 1F1B-1, the performance gains are \textbf{1.28x}, \textbf{1.24x}, and \textbf{1.23x} for 4, 8, and 16 GPUs. For 1F1B-2 they are \textbf{1.13x} and \textbf{1.10x} for 4 and 8 GPUs, whereas the 16 GPU run resulted in an out of memory error. This error is caused by storing the activations and intermediate derivatives of 16 micro-batches on GPU $N-1$. As stated in the previous section, the performance gain should in theory increase with the number of GPUs; however we observe a similar degradation in performance gain, which is most likely due to the increase in communication required.

\subsection{GPU compute occupancy}
As mentioned in section \ref{imp details}, micro-batches are concatenated (copied into contiguous memory location) across their batch dimensions during the compute of backward-p2. In theory this should allow for a greater utilization of compute resources when working with GPUs due to their SIMT architecture. We tested the effect of this concatenation, and the results are summarised in Table \ref{concat_table}. In practice, we did not observe a significant difference in whether the concatenation was performed, or if the $N$ backward-p2 steps were simply performed in a loop (see Figure~\ref{concat b}). It is possible that concatenation does provide speed up, but the concatenation step itself is time consuming and neutralises the benefits.

As the necessary inputs for all backward-p2 operations across the entire model exist in memory at the time of calling the first backward-p2 operation, in theory all backward-p2 operations can be called in parallel. The easiest way to perform this on Nvidia GPUs is via CUDA streams. PyTorch provides an API to use CUDA streams, but we observed that it drastically increased the compute time of the backward-p2 operations, to the point at which it was faster to call each operation in serial. CUDA graphs may be a alternative method to achieve further parallelisation during backward-p2 calls, which we intend to explore in the future. 

\begin{table}[h]
  \centering
  \caption{Average throughput using the 1F1B-1 pipelining schedule and 2BP, with and without concatenating micro-batches during the backward-p2 step.}
  \label{concat_table}
  
  \begin{tabular}{c c c}
   \toprule
   Model & Avg Throughput w/ & Avg Throughput w/o \\
   \midrule
   Transformer-7b& 7120.88 & 7100.69\\
   Bert-Large  & 40427.41  & 40387.41\\
   Mamba-1.4b	& 12437.91  & 12431.13\\
   ResNet152	& 194.93    & 193.10\\
   \bottomrule
  \end{tabular}
\end{table}

\section{Further work}
As stated previously, 2BP can cause an increase in activation memory. Here we propose potential methods to decrease the effects of the increase in activation memory, which we will investigate in the future. Firstly, by performing intermediate derivative checkpointing, which would work similarly to activation checkpointing \cite{act_checkpoint} in which some activations are not stored in memory but recomputed during the backward pass, but applied to the intermediate derivates. Activation checkpointing is a widely used practice for training DNNs. The intermediate derivative recalculations could potentially be overlapped with idle compute to result in minimal performance decreases compared to running 2BP without intermediate derivative checkpointing.

Another potential way to decrease memory consumption caused by storing the intermediate derivatives would be to temporarily store the intermediates from the first few micro-batches into either host memory or NVMe storage \cite{zero_infinity, zero_offload}, and the number of micro-batches to offload would  depend on pipelining schedule and GPU rank.

When training large DNNs, pipeline parallelism is usually not the sole distributed parallelism paradigm employed. Data parallelism is very often used along side pipeline parallelism, and data parallelism can be optimised by overlapping communication of the gradients with computation during the backward propagation through the network. As the gradients are not calculated until the delayed computation during backward-p2, we expect it will be significantly harder to fully overlap communication with computation, especially when working on systems with slower communication infrastructure. We aim to explore this in the future.

\section{Conclusion and discussion}
For this work, we have applied 2BP, which splits the backward propagation step into 2 stages, to training 4 different DNN model architectures. We have demonstrated that this is able to reduce the idle compute time, resulting in significant increase in throughput on top of the SOTA pipelining schedules. 

As part of our implementation of 2BP, the backward pass operations had to be manually implemented; these operations have been implemented by PyTorch but are not exposed to be used functionally though PyTorch's Python API. The authors would like to see this functionality exposed to users, allowing for easier and more granular control over the implementation of custom backward propagation methods such as the one demonstrated in this work. 

\begin{ack}
This work used the Cirrus UK National Tier-2 HPC Service at EPCC (http://www.cirrus.ac.uk) funded by the University of Edinburgh and EPSRC (EP/P020267/1). This work was supported by the Edinburgh International Data Facility (EIDF) and the Data-Driven Innovation Programme at the University of Edinburgh. For the purpose of open access, the author has applied a Creative Commons Attribution (CC BY) licence to any Author Accepted Manuscript version arising from this submission.
\end{ack}

\section*{References}

{
\small
\printbibliography[heading=none]
}

%%%%%%%%%%%%%%%%%%%%%%%%%%%%%%%%%%%%%%%%%%%%%%%%%%%%%%%%%%%%

\appendix

\end{document}